\pdfoutput=1
\documentclass{article}

\PassOptionsToPackage{numbers,sort&compress}{natbib}
\usepackage[preprint]{neurips_2026}

\usepackage[utf8]{inputenc}
\usepackage[T1]{fontenc}
\usepackage{hyperref}
\usepackage{url}
\usepackage{booktabs}
\usepackage{amsfonts}
\usepackage{microtype}
\usepackage{xcolor}
\usepackage{amsmath,amssymb,mathtools}
\usepackage{multirow,array}
\usepackage{graphicx}
\usepackage{placeins}
\usepackage{needspace}
\usepackage{enumitem}
\setlist{itemsep=1pt,topsep=2pt,parsep=0pt}


\newcommand{\zifa}{\textsc{ZifaMem}}
\newcommand{\alps}{\textsc{AlpsBench}}
\newcommand{\vad}{\mathbf{e}}
\newcommand{\dyaffect}{\textsc{DyAffect}}
\newcommand{\clip}{\operatorname{clip}}
\newcommand{\ci}[2]{[#1,#2]}

\title{ZifaMem: Structured Memory for Persona, Preference, and Emotional Continuity in AI Companions}

\author{%
  Jingzhe Fang \\
  ZifaCorp \\
  \texttt{jiawei@zifacorp.com} \\
  \And
  Guozhi Xu \\
  ZifaCorp \\
  \texttt{gavin@zifacorp.com} \\
  \And
  Yunfan Cui \\
  ZifaCorp \\
  \texttt{cuiyunfan@zifacorp.com} \\
  \AND
  Xiaochen Yang\thanks{Xiaochen Yang and Zhangyu Hua did this work during internships at ZifaCorp.} \\
  University of Science and Technology of China \\
  \texttt{yangxiaochen@zifacorp.com} \\
  \And
  Zhangyu Hua\footnotemark[1] \\
  Fudan University \\
  \texttt{huazhangyu@zifacorp.com} \\
}

\begin{document}
\maketitle

\begin{abstract}
AI companions are judged not only by single-turn fluency but by whether they sustain emotional continuity: remembering who the companion is, what the user prefers, and how the relationship has felt. We present \zifa{}, a structured memory system that organizes dialogue into session summaries, episodic memories, and a consolidated user model. Against a deployment-honest comparator that supplies the full raw dialogue history, and under a fixed LLM-as-a-judge protocol with route audits, structured memory raises pooled four-backbone emotional-intelligence scores by $11.4\%$ (95\% CI $6.3\%$ to $17.1\%$), and persona grounding improves on all four backbones (Claude $+42\%$ relative). Multi-turn affect context wins a $+39\%$ net preference over a single-turn snapshot (exploratory), whereas an additional emotion state machine yields no measurable gain on any of five endpoints. Under an identical preregistered protocol, three memory systems---\zifa{}, Mem0, and filtered verbatim retrieval---each improve significantly over raw-history deployment, and \zifa{} and Mem0 are statistically equivalent within $\pm5$ points on the preregistered primary preference endpoint. The \zifa{} SDK, CLI, and portable Agent Skills are open-sourced at \url{https://github.com/zifacorp/zifamem}.
\end{abstract}

\section{Introduction}
\label{sec:intro}

The core experience of an AI companion is emotional continuity. Users describe companion systems through trust, self-disclosure, and shared history \citep{skjuve2021chatbot,brandtzaeg2022myaifriend}; what breaks the experience is rarely a single poorly worded sentence but a discontinuity---a companion that forgets last week's grief, contradicts its own persona, or repeats a preference the user has already corrected. Memory is the carrier of that continuity: a compact, organized representation of the relationship can surface the persona, preferences, and emotional history that a backbone may fail to extract from a long raw transcript. Long-term-memory research, by contrast, has focused primarily on factual recall, temporal reasoning, and profile consistency \citep{packer2023memgpt,zhong2023memorybank,chhikara2025mem0,maharana2024locomo,wu2025longmemeval}; this paper asks what structured memory does for the relational and emotional workloads that define companionship.

A second intuition is that believable affect should have temporal structure. Psychological models describe core affect through a home base, variability, and attractor strength, and identify emotional inertia as persistence in affective state \citep{kuppens2010dynaffect,kuppens2010inertia}. Computational emotion systems therefore often combine appraisal with a state update rather than classifying each turn independently \citep{ortony1988occ,marsella2009ema}. For an LLM companion, however, the engineering plausibility of a state variable does not establish its response value. A memory context may already carry the history from which affect is inferred, and a strong backbone may integrate recent emotional cues without an explicit dynamical system.

This paper studies memory-carried emotional continuity through five linked evaluations. First, we measure \zifa{} as a structured context representation against the deliberately strong comparator of full raw dialogue history, across four backbones and four capabilities. Second, we analyze a paired $2\times2$ matrix that adds or removes the legacy affect layer within both context policies. Third, we introduce \dyaffect{}, a confidence-adaptive state estimator that separates user affect from companion self-state, and evaluate it against both a current-turn snapshot and a confidence-weighted three-observation baseline. Fourth, a preregistered same-protocol experiment compares \zifa{} with prior memory systems---Mem0 and a filtered verbatim-retrieval baseline---on the best-powered endpoints. Fifth, a preregistered layered study measures how the memory advantage changes as history length grows from 16k to over 100k tokens. Three of these evaluations---the \dyaffect{} contrastive suite and the prior-system and long-history experiments---were preregistered, with stopping rules frozen before data; the factorial matrix and capability-map analyses are exploratory, route-audited runs with per-row artifacts (Appendix~\ref{app:prereg-ledger}). In every stream, a failed primary comparison is reported as a failure and is not replaced by a favorable secondary result.

The evidence supports a targeted thesis rather than a universal one. The most general measured gain is emotional continuity: the structured representation improves judged emotional intelligence directionally on every verified backbone. Capability-level gains form a map rather than a constant: they concentrate where a backbone leaves headroom on persona and preference workloads. At the deployment level, replacing raw history with a memory system helps consistently; at the representation level, \zifa{} is statistically equivalent to an existing memory system within $\pm5$ percentage points on the preregistered primary endpoint (post-hoc TOST), differences on emotional intelligence are not resolved at current power, and filtered verbatim retrieval is a strong baseline on short-history preference tasks. The tested affect layer can help when structured memory is absent, but its sign depends on the backbone, and its incremental contribution under memory is undetectable on the best-powered endpoints. Affect information and affect dynamics are therefore separate design choices.

Our contributions are:
\begin{itemize}
  \item a routing-verified evaluation that identifies emotional continuity as the most general structured-memory gain, together with a four-backbone $\times$ four-capability effect map that localizes persona and preference gains where baseline headroom exists;
  \item a preregistered, same-protocol comparison with prior memory systems whose honest verdict is primary-endpoint equivalence within $\pm5$ percentage points (post-hoc TOST) with emotional-intelligence differences unresolved at $n=52$---and the finding that simple filtered verbatim retrieval is an undervalued strong baseline for short-history preference workloads;
  \item a preregistered long-history layered evaluation showing a directionally positive but underpowered advantage gradient, with an explicit power statement;
  \item a paired decomposition showing where affect prompting adds information, that its sign without memory is backbone-dependent, and that it is uniformly null once memory is present;
  \item \dyaffect{}, a confidence-adaptive, change-point-aware estimator of user-affect state and trend, with a contrastive protocol separating the value of a current snapshot, recent affect history, and latent temporal state; and
  \item a falsification record spanning response quality, deterministic mechanics, and component-level implementation provenance.
\end{itemize}

\section{System Decomposition and Provenance}
\label{sec:system}

\begin{figure}[h!]
  \centering
  \includegraphics[width=\linewidth]{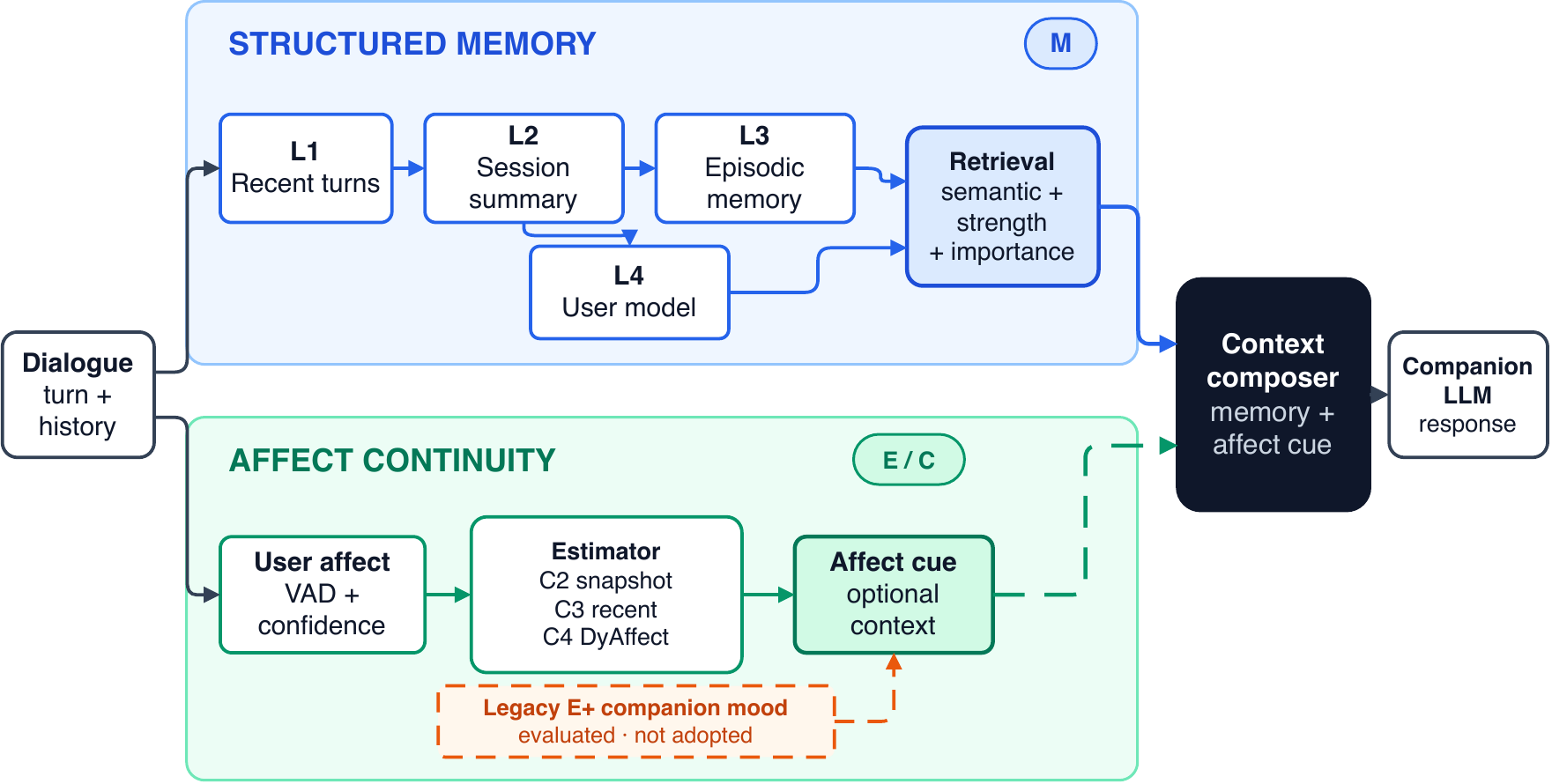}
  \caption{System decomposition of structured memory and affect continuity.
  Dialogue branches into the structured-memory and affect-continuity lanes,
  which converge at the context composer. The memory lane contains L1--L4
  consolidation and retrieval; the affect lane compares C2, C3, and C4 while
  retaining the historical $E+$ companion-mood block as an evaluated but unadopted
  path.}
  \label{fig:system-framework}
\end{figure}

Figure~\ref{fig:system-framework} separates the two sources of response context
evaluated in this work. The structured-memory lane transforms dialogue into
layered relationship memory and query-time retrieval; affect tags participate
in both consolidation and retrieval, so this lane---not a separate emotion
module---is what carries emotional history into the response. The
affect-continuity lane exposes current or recent user-affect observations, a
latent \dyaffect{} estimate, or the separately evaluated legacy companion-mood path.
The lanes meet only at response composition, which keeps memory effects
distinct from affect-prompt and temporal-state effects.

\FloatBarrier

\subsection{Affect-tagged structured memory substrate}
\label{sec:memory}

\zifa{} organizes relationship context into a rolling turn buffer, session summaries, episodic memories, and a consolidated user model. Emotional history is not stored in a separate affect module: it is tagged into the memory representation itself. Session summaries promote to episodic memory when either their importance $I$ reaches the runtime threshold or an emotional peak was recorded:
\begin{equation}
\label{eq:gate}
\operatorname{promote}(s)=\mathbf{1}\!\left[I_s\geq0.5\;\lor\;\operatorname{had\_emotional\_peak}(s)\right].
\end{equation}
This Boolean gate corrects an earlier manuscript description that used an unsupported continuous-intensity threshold. Episodic strength is recomputed at retrieval from age, access count, importance, and stored affect metadata. The evaluated reranker then uses
\begin{equation}
\label{eq:retrieval}
\operatorname{score}(q,j)=0.35\,\operatorname{sem}(q,j)+0.30\,s_j+0.35\,I_j.
\end{equation}
There is no separately weighted $0.10$ emotional-intensity term in the evaluated path. Affect can still influence which memories survive through the promotion gate and strength calculation, but the retrieval equation should not double-count a term the implementation did not use. The result is that emotionally salient episodes are preferentially consolidated and retrieved, which is the design hypothesis behind the emotional-continuity results of Section~\ref{sec:results}.

The public repository provides an alpha, dependency-light SDK for this memory substrate. The response evaluation predates the public SDK and was produced by an internal research system sharing the same memory concepts. We therefore treat the public SDK, the evaluated research system, and the current deployed service as distinct implementation layers.

\subsection{Legacy affect observation and injection}
\label{sec:legacy}

The legacy affect path appraises how the user's utterance affects the companion, updates a companion VAD state $\vad(t)\in[-1,1]^3$, maps the state to an emotion label, and injects a prompt block containing the label, intensity, candidate labels, and personality style hint. In the paired experiment, $E+$ denotes this complete path and $E-$ removes it. The injected quantity is therefore the companion's appraised mood, not a validated label of the user's own emotion.

The conceptual dynamics use a driven second-order system,
\begin{equation}
\label{eq:dynamics}
m\ddot{\vad}(t)=\sum_i\mathbf{f}_i(t)-k\bigl(\vad(t)-\vad_0\bigr)-c\dot{\vad}(t).
\end{equation}
For one component with no force, $x=e-e_0$ satisfies $mx''+cx'+kx=0$. In the overdamped case its general solution is
\begin{equation}
\label{eq:roots}
x(t)=A e^{r_+t}+B e^{r_-t},\qquad
r_{\pm}=\frac{-c\pm\sqrt{c^2-4mk}}{2m}.
\end{equation}
The earlier single-exponential expression is not the general closed form of Eq.~\eqref{eq:dynamics}; it also discards velocity. This distinction matters because the evaluated fast-forward implementation used that approximation.

A frozen legacy memory-feedback mechanism computed an importance-weighted absolute VAD centroid, multiplied it by mean importance, and applied the result as an eight-second open-loop force with componentwise saturation. It did not subtract the current VAD and therefore was not a closed-loop attractor. This mechanism is evaluated only as a historical design and is not presented as part of the current deployed system.

\begin{figure}[t]
  \centering
  \includegraphics[width=\linewidth]{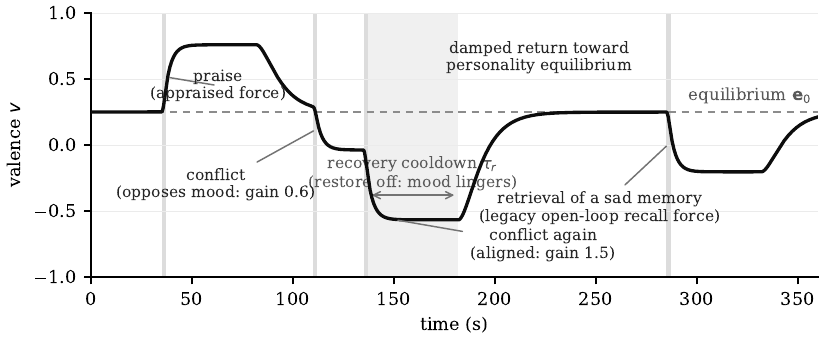}
  \caption{Legacy affect dynamics under a synthetic event sequence. The audited numerical integrator illustrates appraisal forces, inertia-dependent gains, a recovery cooldown, damped return toward a personality equilibrium, and an open-loop recall force. This is an explanatory mechanics figure, not evidence of improved response quality; Section~\ref{sec:mechanical-results} reports where the implementation departs from its intended behavior.}
  \label{fig:legacy-dynamics}
\end{figure}
\FloatBarrier

\Needspace{0.38\textheight}
\subsection{\dyaffect{}: Confidence-Adaptive User-Affect State Estimation}
\label{sec:dyaffect-method}

\dyaffect{} frames emotional continuity as online state estimation under noisy, confidence-labeled observations. Each turn produces a user-affect observation $o_t=(\mathbf{v}_t,c_t)$ with VAD value $\mathbf{v}_t$, confidence $c_t$, and an optional explicit-transition flag. From the previous state $\mathbf{s}_{t-1}$ and trend $\mathbf{h}_{t-1}$, it forms a bounded prediction $\mathbf{p}_t$ and innovation $\boldsymbol{\epsilon}_t$:
\begin{align}
\label{eq:dyaffect-predict}
\mathbf{p}_t &= \clip_{[-1,1]^3}\!\left(\mathbf{s}_{t-1}+\omega_t\mathbf{h}_{t-1}\right),
& \boldsymbol{\epsilon}_t &= \mathbf{v}_t-\mathbf{p}_t, \\
\label{eq:dyaffect-update}
g_t &= \clip_{[g_{\min},g_{\max}]}\!\left(g_{\min}+\beta c_t+\gamma\chi_t+\delta\rho_t\right),
& \mathbf{s}_t &= \clip_{[-1,1]^3}\!\left(\mathbf{p}_t+g_t\boldsymbol{\epsilon}_t\right).
\end{align}
Here $\chi_t$ activates on an explicit or high-confidence change point, and $\rho_t$ activates on a confidence-gated valence reversal. The trend update
\begin{equation}
\label{eq:dyaffect-trend}
\mathbf{h}_t=\eta\mathbf{h}_{t-1}+(1-\eta)(\mathbf{s}_t-\mathbf{s}_{t-1})
\end{equation}
retains gradual motion while the adaptive gain lets explicit reversals override stale momentum. Recovery observations can increase $\omega_t$, allowing a trajectory to continue toward neutral instead of freezing at the most recent negative observation.

The contrastive design isolates which part of this machinery contributes response value:
\begin{itemize}
  \item \textbf{C2, current snapshot}: expose only $o_t$;
  \item \textbf{C3, recent observations}: use the confidence-weighted mean of the latest three observations, with no latent state; and
  \item \textbf{C4, \dyaffect{}}: maintain state and trend with confidence-adaptive gain, explicit change-point handling, and bounded prediction.
\end{itemize}
\dyaffect{} also defines an orthogonal companion self-state for stylistic control. This separation invariant prevents the agent's own mood from rewriting beliefs about the user: user-affect state governs acknowledgment, certainty, transition framing, and response goal, while companion self-state may influence character style only. The response experiment compares the current snapshot, recent observations, and \dyaffect{}; user-affect/companion-self-state separation is an architectural invariant rather than an additional experimental condition. To isolate the state-estimation hypothesis, \dyaffect{} is evaluated independently of deployment-specific prompting and service orchestration.

\section{Benchmark and Experimental Protocol}
\label{sec:protocol}

\subsection{AlpsBench endpoints and factorial cells}

\alps{} \citep{xiao2026alpsbench} is an LLM-personalization benchmark of long-term interaction sequences built from real human--LLM dialogues curated from WildChat \citep{zhao2024wildchat}, paired with human-verified structured memories; dialogues are bilingual (Chinese/English) and distributed under the benchmark's released data-usage terms, with test-set gold withheld by its maintainers. We use its public Task~4 dev split throughout. Task 4 replays a dialogue and asks the backbone to answer a probe. We report A1 persona grounding ($n=114$), A2 content and interaction-style preference following ($n=246$ each), A3 fiction--reality discrimination ($n=100$), and A5 emotional intelligence ($n=52$, score 1--5). A1--A3 are binary per instance; A5 is averaged over rubric dimensions. A4 was not executed. We operationalize the \emph{retrieval side} of emotional continuity as A5: whether remembered emotional context is recognized and engaged in the response. The temporal side of the construct---trajectory naturalness and cross-session affective consistency---is not measured by this endpoint, and no claim in this paper covers it.

The paired matrix has four cells: raw history without affect ($M-E-$), raw history with affect ($M-E+$), structured memory without affect ($M+E-$), and structured memory with affect ($M+E+$). The notation $M-$ is retained for compatibility, but it is not empty context: it supplies the full raw dialogue history. Consequently, $M|E-$ is a system comparison between two context representations. By contrast, $E|M-$ and $E|M+$ are clean within-layer simple effects because the two sides of each contrast share the same context policy.

\subsection{Qualification, judging, and uncertainty}
\label{sec:qualification}

We include only responses whose serving records unambiguously identify the generating model. An earlier Gemini factorial cohort is excluded because one condition was inadvertently served by Claude and the remaining skipped responses lacked sufficient provenance to reconstruct their routes. A separately routed Gemini cohort remains admissible for the structured-memory comparison. Fresh, explicitly routed Gemini pairs are now admissible for both within-layer simple effects: $E|M+$ from an independent replication and $E|M-$ from a same-day, same-batch pair. The Gemini $M|E-$ contrast crosses two batches run three days apart under the same endpoint, judge, and protocol; it is admissible only with that batch separation disclosed. These exclusions and qualifications are applied by cohort, not by model family.

Responses are re-judged by a fixed DeepSeek-v4 Ark endpoint with thinking disabled, temperature zero, and three repeats. Binary endpoints use majority vote; A5 uses the mean. We compute paired instance bootstrap intervals with 10,000 samples. Reported sign-test $p$-values in the supplementary audit are exploratory and uncorrected, so the paper emphasizes estimates and intervals. A separate 4,096-row bridge compares these judgments with the earlier DeepSeek-v3.1 judge (Appendix~\ref{app:judge}).

\subsection{DyAffect contrastive response test}

The authored development suite contains 96 scenarios spanning eight affect-transition families, three intensities, two languages, and two paraphrases. For each scenario, a common base response is edited only when two estimators produce different response plans. C4 differs from C3 on 60 scenarios and from C2 on 38. Each changed pair is judged in both answer orders with condition labels hidden. A consistent win or loss requires both orders to agree; discordant outcomes are neutral in the net-preference numerator and are reported separately.

The frozen primary gate requires C4 to beat C3 by at least $+0.10$ net preference with a bootstrap lower bound above $-0.05$, plus naturalness and order-stability gates. All gates are conjunctive. The same development evidence was used to iterate the estimator, so even a pass would have authorized only a new confirmation design, not a final claim.

\subsection{Preregistered prior-system comparison}
\label{sec:exp1-protocol}

The design of the prior-memory-system experiment was fixed by a preregistration frozen before any data collection, \texttt{positive\_exp\_baselines\_prereg\_20260714}. All conditions receive the same replayed inputs; the \texttt{[Relevant Memories]} block position and the Task~4 system prompt are byte-identical across systems by assertion. The generating backbone is Claude-Sonnet-4.6 on an explicit ZenMux route, with row-level route audits over all $298+298$ new rows reporting zero anomalies, and the judge is the same fixed Ark endpoint with three repeats and thinking disabled. The comparison systems are Mem0 (\texttt{mem0ai} 2.0.12 with a Doubao LLM, \texttt{doubao-embedding-vision}, and a local Qdrant store) and a preregistered additional column, simple filtered verbatim retrieval (``simple RAG''): ten-turn dialogue chunks, top-5 by the same embedding model. The \zifa{} and raw-history columns reuse the pilot cohorts run five to six days earlier under the same endpoint, judge, and prompt; this cross-batch reuse is disclosed wherever those columns are cited. Judge errors were zero; one empty Mem0 A2 answer is scored as a failure per the frozen rule. Endpoints are A2 preference following ($n=246$ paired) and A5 emotional intelligence ($n=52$ paired), with paired bootstrap (10,000 resamples, seed 20260715). The preregistered primary endpoint is A2 interaction-style preference, contrast \zifa{} minus Mem0.

\subsection{Preregistered long-history layered protocol}
\label{sec:exp2-protocol}

A second preregistration (\texttt{positive\_exp\_longhistory\_prereg\_20260714}) governs the history-length experiment. Because the natural length distribution of LongMemEval-s has degenerated to roughly $120$k tokens $\pm2\%$, the preregistration constructs nested history layers: L1 targets 16k tokens, L2 64k, and L3 the full history, with evidence sessions always preserved and filler sessions removed by seeded, order-preserving uniform subsampling. We sample $n=48$ questions stratified by question type. Conditions are raw-truncated (the most recent 32k tokens of raw messages) versus the full \zifa{} pipeline replayed over the layer with memory injection. The backbone is Claude-Sonnet-4.6 on an explicit ZenMux route, judged by the same Ark endpoint with three repeats. All $144/144$ cells completed with zero cell errors and zero route-audit anomalies. Scores are on a 0--5 scale (mean of three judge repeats) with paired bootstrap (10,000 resamples, seed 20260716). The preregistered primary hypothesis H1 is $\Delta_{\mathrm{L3}}>0$; H2 is the gradient $\Delta_{\mathrm{L3}}-\Delta_{\mathrm{L1}}>0$.

\FloatBarrier
\section{Main Results: Where Structured Memory Is Strong}
\label{sec:results}

\subsection{Emotional continuity replicates across backbones}
\label{sec:a5-results}

The most general measured gain is emotional continuity. Table~\ref{tab:a5-memory} reports the A5 structured-context effect against full raw history across the four backbones with verified model routes. Claude and Qwen3.6 have individually positive intervals ($+0.64$ and $+0.56$); Doubao-Seed-1.8 and GLM-5.2 are positive but uncertain. Pooling the four paired sets gives $+0.42$ \ci{+0.25}{+0.60}. Relative to the pooled raw-history baseline of $3.70$, this is an $11.4\%$ improvement (95\% CI $6.3\%$ to $17.1\%$, paired ratio bootstrap, seed 20260717). This result establishes that the affect-tagged structured representation can improve judged emotional intelligence without withholding the raw-history comparator from the baseline: the baseline sees everything the memory system saw, in raw form.

\begin{table}[htbp]
\centering
\caption{A5 structured-context effect in the $E-$ layer across four backbones with verified model routes. The contaminated historical Gemini factorial cohort is excluded; the admissible fresh Gemini factorial pair is reported in Table~\ref{tab:gemini}.}
\label{tab:a5-memory}
\small
\begin{tabular}{lrr}
\toprule
Backbone & $M|E-$ & 95\% CI \\
\midrule
Claude-Sonnet-4.6 & +0.64 & \ci{+0.27}{+1.01} \\
Qwen3.6 & +0.56 & \ci{+0.17}{+0.97} \\
Doubao-Seed-1.8 & +0.23 & \ci{-0.11}{+0.58} \\
GLM-5.2 & +0.27 & \ci{-0.03}{+0.58} \\
\midrule
Pooled ($n=208$) & +0.42 & \ci{+0.25}{+0.60} \\
\bottomrule
\end{tabular}
\end{table}

\subsection{A capability-level applicability map}
\label{sec:capability-map}

Table~\ref{tab:claude-memory} gives the cleanest high-powered comparison for the target regime. Relative to full raw history, Claude gains $0.237$ on persona grounding and $0.142/0.150$ on the two preference endpoints. These are precisely the companion workloads for which an organized user model should reduce repeated inference from a long transcript. A3 is already at ceiling and changes little. A5 rises by $0.635$, although its $n=52$ interval is wider.

\begin{table}[!htb]
\centering
\caption{Claude system comparison in the $E-$ layer. $\Delta$ is ZifaMem context minus full raw dialogue history; it is not an empty-context memory main effect. Intervals use paired instance bootstraps.}
\label{tab:claude-memory}
\small
\setlength{\tabcolsep}{4pt}
\begin{tabular}{lrrrr}
\toprule
Endpoint & $n$ & Raw history & ZifaMem & $\Delta$ (95\% CI) \\
\midrule
A1 persona & 114 & .561 & .798 & +.237 \ci{+.140}{+.333} \\
A2 preference (content) & 246 & .492 & .634 & +.142 \ci{+.077}{+.207} \\
A2 preference (style) & 246 & .512 & .663 & +.150 \ci{+.089}{+.211} \\
A3 fiction--reality & 100 & .960 & .980 & +.020 \ci{-.020}{+.060} \\
A5 emotional intelligence & 52 & 3.628 & 4.263 & +.635 \ci{+.269}{+1.003} \\
\bottomrule
\end{tabular}
\end{table}

Table~\ref{tab:m-matrix} extends the capability comparison to the full four-backbone $\times$ four-capability matrix, all cells route-audited under the DeepSeek-v4 judge. The capability-level gains are highly concentrated in Claude, whose A1/A2 contrasts are the only individually significant cells besides Doubao's A2 content endpoint. Qwen3.6 and GLM-5.2 show no A2 gain or a slight regression---and their raw-history A2 baselines are all higher than Claude's, which is the capability-level version of the headroom pattern: structured memory helps most where the backbone leaves room on the workload. A1 persona grounding is directionally positive on all four backbones. Together with Table~\ref{tab:a5-memory}, the matrix supports a targeted-applicability reading: capability-level benefit depends on the backbone--workload combination, while the emotional-intelligence gain is the most general.

\begin{table}[htbp]
\centering
\caption{Structured-context effects (\zifa{} minus full raw history) across four backbones and four capability endpoints, DeepSeek-v4 judge with three repeats, paired instance bootstraps. Endpoint definitions and instance counts follow Table~\ref{tab:claude-memory}. The Claude rows restate the Table~\ref{tab:claude-memory} contrasts under the 2026-07-14 canonical recomputation; interval endpoints can differ in the third decimal across bootstrap seeds.}
\label{tab:m-matrix}
\small
\setlength{\tabcolsep}{4pt}
\begin{tabular}{llrrr}
\toprule
Backbone & Endpoint & Raw history & ZifaMem & $\Delta$ (95\% CI) \\
\midrule
Claude-Sonnet-4.6 & A1 persona & .561 & .798 & +.237 \ci{+.140}{+.333} \\
 & A2 preference (content) & .492 & .634 & +.142 \ci{+.081}{+.207} \\
 & A2 preference (style) & .512 & .663 & +.150 \ci{+.089}{+.211} \\
 & A3 fiction--reality & .960 & .980 & +.020 \ci{-.020}{+.060} \\
\midrule
Qwen3.6 & A1 persona & .702 & .781 & +.079 \ci{-.009}{+.167} \\
 & A2 preference (content) & .589 & .569 & -.020 \ci{-.085}{+.045} \\
 & A2 preference (style) & .614 & .602 & -.012 \ci{-.077}{+.053} \\
 & A3 fiction--reality & .760 & .710 & -.050 \ci{-.150}{+.060} \\
\midrule
GLM-5.2 & A1 persona & .702 & .772 & +.070 \ci{-.018}{+.158} \\
 & A2 preference (content) & .663 & .606 & -.057 \ci{-.118}{+.004} \\
 & A2 preference (style) & .667 & .634 & -.033 \ci{-.093}{+.028} \\
 & A3 fiction--reality & .800 & .820 & +.020 \ci{-.060}{+.100} \\
\midrule
Doubao-Seed-1.8 & A1 persona & .711 & .772 & +.061 \ci{-.035}{+.158} \\
 & A2 preference (content) & .573 & .642 & +.069 \ci{+.004}{+.134} \\
 & A2 preference (style) & .614 & .659 & +.045 \ci{-.020}{+.110} \\
 & A3 fiction--reality & .180 & .140 & -.040 \ci{-.110}{+.030} \\
\bottomrule
\end{tabular}
\end{table}

\begin{figure}[tp]
  \centering
  \includegraphics[width=\linewidth]{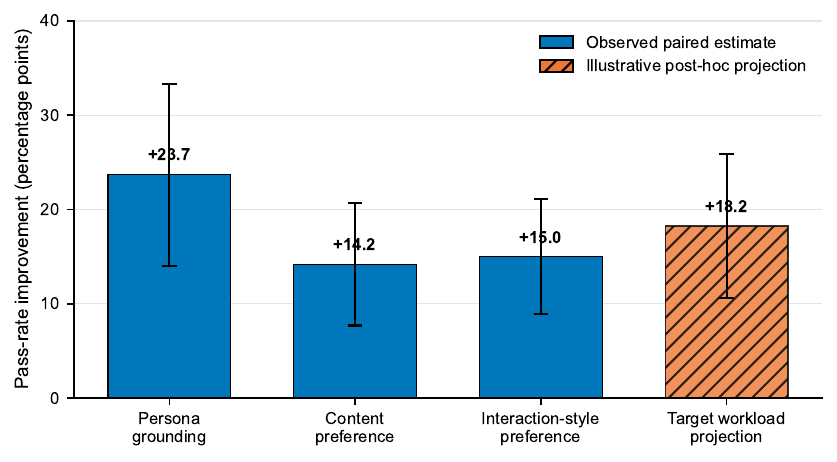}
  \caption{Observed Claude pass-rate gains and an illustrative target-workload projection. Solid bars are paired benchmark estimates with 95\% confidence intervals. The hatched bar combines A1 persona ($40\%$), A2 content preference ($30\%$), and A2 interaction-style preference ($30\%$). Its whisker is a weighted envelope of marginal interval endpoints, not a joint confidence interval. The target profile is post-hoc and is not an observed benchmark cell.}
  \label{fig:targeted-advantage}
\end{figure}

Figure~\ref{fig:targeted-advantage} converts the three binary Claude target endpoints into one illustrative application profile: $40\%$ persona grounding, $30\%$ content preference, and $30\%$ interaction-style preference. Applying those weights to the observed paired deltas gives a projected $+18.2$ percentage-point pass-rate improvement. The displayed $+10.6$ to $+25.9$ envelope applies the same weights to the three marginal interval endpoints; it is not a joint confidence interval. The profile and weights were chosen after outcome review, so this is a post-hoc deployment illustration rather than a new benchmark result, and Table~\ref{tab:m-matrix} shows it would not transfer to backbones without the corresponding headroom.

\FloatBarrier

\subsection{Emotional threads beat snapshots; affect injection is backbone-dependent}
\label{sec:threads}

Two independent evidence streams indicate that what helps is the emotional \emph{thread}---multi-turn affective history---rather than a single injected affect signal.

First, the \dyaffect{} contrastive test provides a direct comparison between multi-turn affect context and a single-turn snapshot. C4 beats the current-snapshot baseline C2 by $+0.395$ \ci{+0.211}{+0.579} on the 38 scenarios where their plans differ (Table~\ref{tab:dyaffect}). This estimate carries its preregistered caveats verbatim: it is exploratory, its answer-order discordance is $18.4\%$, above the frozen $15\%$ limit, and C3 already supplies multi-turn observations and beats C4 on the primary contrast (Section~\ref{sec:dyaffect-results}). What survives those caveats is directional: multi-turn affect context can be more useful than a single appraisal, which is exactly the information that affect-tagged memory is designed to carry. Mechanism-level attribution of the A5 gain to the affect tags themselves is not isolated by these experiments: the knockout runs that would remove affect tags from consolidation or retrieval were not executed, so the measured gain may derive from the structured representation as a whole; isolating the affect-tag contribution remains future work.

\begin{table}[!htb]
\centering
\caption{\dyaffect{} contrastive response discovery. Net preference is computed on changed-plan scenarios requiring agreement across both answer orders. The failed primary comparison is analyzed in Section~\ref{sec:dyaffect-results}.}
\label{tab:dyaffect}
\small
\begin{tabular}{lrrrrl}
\toprule
Comparison & Changed & W/L/N & Net & 95\% CI & Status \\
\midrule
C4 vs C3 recent observations & 60 & 12/20/28 & -.133 & \ci{-.317}{+.050} & Primary fail \\
C4 vs C2 current snapshot & 38 & 17/2/19 & +.395 & \ci{+.211}{+.579} & Exploratory \\
\bottomrule
\end{tabular}
\end{table}

\begin{figure}[t]
  \centering
  \includegraphics[width=0.82\linewidth]{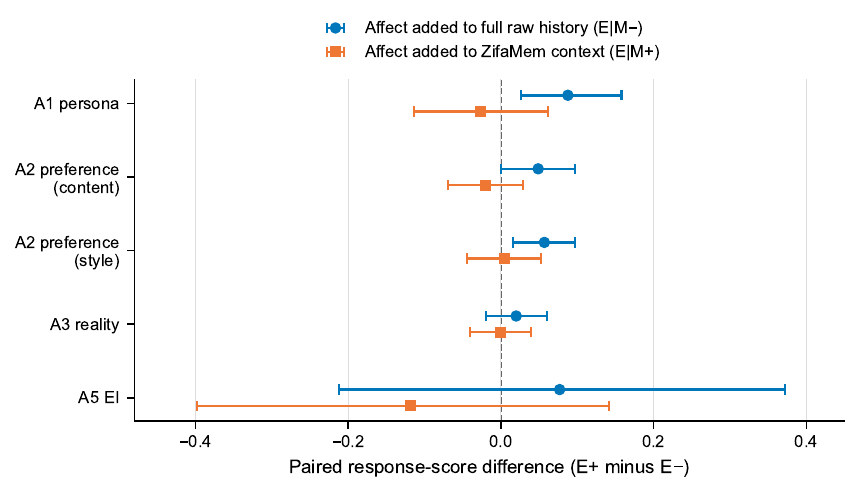}
  \caption{Paired simple effects of adding the legacy affect layer to raw-history and ZifaMem context policies on Claude. Points are means and bars are 95\% paired-bootstrap intervals. The pattern is consistent with partial substitution, but the interaction mechanism is not identified, and the raw-history-side benefit is backbone-dependent (Section~\ref{sec:factorial}).}
  \label{fig:affect-simple}
\end{figure}

Second, the paired factorial matrix shows that injecting an affect block is not a reliable substitute for that thread. Table~\ref{tab:affect-simple} and Figure~\ref{fig:affect-simple} report the two clean within-layer simple effects on Claude. When the context is full raw history, adding the legacy affect block improves A1 by $0.088$ and A2 interaction-style preference following by $0.057$, with positive intervals; A2 content preference is positive at $0.049$ with a lower bound at zero. But this benefit does not generalize across backbones: on a fresh, explicitly routed, same-batch Gemini A5 pair, the same contrast is significantly \emph{negative}, $E|M-=-0.346$ \ci{-0.696}{-0.032}. Without structured memory, the sign of affect injection depends on the backbone---positive on Claude, negative on Gemini. The only cross-backbone invariant in this design is the null under memory (Section~\ref{sec:factorial}). Affect injection is therefore not a dependable channel for emotional continuity; the memory-carried thread is the reliable one.

\begin{table}[!ht]
\centering
\caption{Claude simple effects of the legacy affect layer. Each contrast is paired within one context policy. The $E|M-$ column is backbone-specific: the fresh same-batch Gemini A5 pair yields $-0.346$ \ci{-0.696}{-0.032} (Section~\ref{sec:factorial}).}
\label{tab:affect-simple}
\small
\setlength{\tabcolsep}{4pt}
\begin{tabular}{lrrr}
\toprule
Endpoint & $n$ & $E|M-$ (95\% CI) & $E|M+$ (95\% CI) \\
\midrule
A1 persona & 114 & +.088 \ci{+.026}{+.158} & -.026 \ci{-.114}{+.061} \\
A2 preference (content) & 246 & +.049 \ci{.000}{+.098} & -.020 \ci{-.069}{+.029} \\
A2 preference (style) & 246 & +.057 \ci{+.016}{+.098} & +.004 \ci{-.045}{+.053} \\
A3 fiction--reality & 100 & +.020 \ci{-.020}{+.060} & .000 \ci{-.040}{+.040} \\
A5 emotional intelligence & 52 & +.077 \ci{-.212}{+.372} & -.119 \ci{-.397}{+.141} \\
\bottomrule
\end{tabular}
\end{table}
\FloatBarrier

\FloatBarrier

\section{Boundaries and System Selection}
\label{sec:boundaries}

This section discloses, compactly and completely, where the evidence bounds the claims: how \zifa{} compares with prior memory systems under an identical protocol, how the advantage behaves as history grows, which backbones define boundary cases, and what the affect state machine does and does not add.

\subsection{Prior memory systems: primary-endpoint equivalence, and a strong retrieval baseline}
\label{sec:systems-results}

Table~\ref{tab:systems} reports the preregistered same-protocol comparison (Section~\ref{sec:exp1-protocol}). The preregistered primary endpoint verdict is unchanged from the frozen analysis output: on A2 interaction-style preference, \zifa{} $-$ Mem0 $= +0.004$ \ci{-0.049}{+0.057} $\rightarrow$ \textbf{null}. Because a confidence interval containing zero shows only absence of a detected difference, we additionally ran a post-hoc TOST equivalence check on the same per-instance artifacts: the 90\% paired-bootstrap CI of the primary contrast, $[-0.041,+0.049]$, lies inside a $\pm0.05$ equivalence margin, so \zifa{} and Mem0 are statistically equivalent within five percentage points on the primary endpoint. The margin was chosen post hoc, matching the judge-noise band (Appendix~\ref{app:judge}). On A2 content preference the 90\% CI, $[-0.053,+0.041]$, marginally exceeds the margin, so we claim only no detectable difference there. At the deployment level, all three memory systems improve significantly over raw history on every endpoint tested. On A5 emotional intelligence, both paired contrasts against \zifa{} cross zero at the 95\% level, but the intervals are wide (\zifa{} $-$ Mem0 $= -0.199$ \ci{-0.445}{+0.029}) and far exceed any equivalence margin: representation-level differences on emotional intelligence are \emph{not resolved} at $n=52$, and no parity claim is made on this endpoint. On the short-history A2 preference endpoints \zifa{} trails simple filtered verbatim retrieval by a significant $-0.110$ \ci{-0.163}{-0.057} on both endpoints.

\begin{table}[htbp]
\centering
\caption{Preregistered same-protocol comparison with prior memory systems on Claude-Sonnet-4.6 (A2 $n=246$ paired, A5 $n=52$ paired; paired bootstrap, 10{,}000 resamples, seed 20260715). The \zifa{} and raw-history columns reuse pilot cohorts run five to six days earlier under the same endpoint, judge, and prompt (disclosed in Section~\ref{sec:exp1-protocol}). The primary endpoint is A2 style, \zifa{} $-$ Mem0. Post-hoc TOST (90\% paired-bootstrap CI vs.\ a $\pm0.05$ margin chosen post hoc, matching the judge-noise band): A2 style $[-0.041,+0.049]$, within the margin $\rightarrow$ equivalent within $\pm5$pp; A2 content $[-0.053,+0.041]$ and A5 $[-0.401,-0.006]$ exceed the margin $\rightarrow$ no equivalence claimed; A5 differences are unresolved at $n=52$.}
\label{tab:systems}
\small
\setlength{\tabcolsep}{4pt}
\begin{tabular}{lrrr}
\toprule
 & A2 preference (content) & A2 preference (style) & A5 emotional intelligence \\
\midrule
Raw history & .492 & .512 & 3.628 \\
Mem0 & .638 & .659 & 4.462 \\
Simple RAG & .744 & .772 & 4.138 \\
\zifa{} & .634 & .663 & 4.263 \\
\midrule
Mem0 $-$ raw & +.146 \ci{+.081}{+.215} & +.146 \ci{+.081}{+.211} & +.833 \ci{+.404}{+1.285} \\
RAG $-$ raw & +.252 \ci{+.195}{+.309} & +.260 \ci{+.203}{+.317} & +.510 \ci{+.099}{+.926} \\
\zifa{} $-$ Mem0 & -.004 \ci{-.061}{+.049} & \textbf{+.004 \ci{-.049}{+.057}} & -.199 \ci{-.445}{+.029} \\
\zifa{} $-$ RAG & -.110 \ci{-.163}{-.057} & -.110 \ci{-.163}{-.057} & +.125 \ci{-.144}{+.401} \\
\bottomrule
\end{tabular}
\end{table}

We claim the second-order result as a finding: \textbf{simple filtered verbatim retrieval is an undervalued strong baseline} for short-history preference workloads. When the probe asks the model to honor a stated preference and the evidence fits inside a few retrieved verbatim chunks, retrieval preserves exact wording that summarization abstracts away, and it does so with far less machinery. The honest implication for positioning is that the paper's supported claim is deployment-level---replacing raw-history context with a memory system helps, and on the preregistered primary endpoint \zifa{} delivers that benefit at a level statistically equivalent to an existing open system---while representation-level superiority over prior systems is not supported by these endpoints and is not claimed. On A5, the workload closest to this paper's thesis, the contrasts among the structured and retrieval representations are not resolved at $n=52$; distinguishing them is a power problem we leave explicitly open.

\subsection{The long-history gradient is directionally positive but unproven}
\label{sec:longhistory-results}

Table~\ref{tab:longhistory} reports the preregistered layered experiment (Section~\ref{sec:exp2-protocol}). The preregistered primary verdict, under the frozen stopping rule, is that \textbf{H1 is not supported}: at the full-history layer L3, $\Delta = +0.118$ \ci{-0.660}{+0.889}, an interval that crosses zero. The gradient across layers moves in the mechanistically expected direction---$-0.30$ at 16k, $+0.17$ at 64k, $+0.12$ at 104k, with H2 $(\Delta_{\mathrm{L3}}-\Delta_{\mathrm{L1}}) = +0.417$ \ci{-0.438}{+1.312}---but at $n=48$ per layer the noise dominates. Detecting effects of the observed $0.1$--$0.4$ magnitude would require roughly $n\gtrsim150$ per layer or a binarized endpoint; we record this as a power statement, not a positive result.

\begin{table}[htbp]
\centering
\caption{Preregistered long-history layered comparison on Claude-Sonnet-4.6 ($n=48$ per layer; score 0--5 is the mean of three judge repeats; paired bootstrap, 10{,}000 resamples, seed 20260716). Truncated = most recent 32k tokens of raw messages. H1 is the L3 row; H2 is the L3$-$L1 gradient.}
\label{tab:longhistory}
\small
\setlength{\tabcolsep}{4pt}
\begin{tabular}{lrrrr}
\toprule
Layer (median tokens) & Truncated & \zifa{} & $\Delta$ (95\% CI) & Verdict \\
\midrule
L1 ($\sim$14.6k) & 4.271 & 3.972 & -.299 \ci{-.757}{+.167} & \\
L2 ($\sim$62.9k) & 3.229 & 3.396 & +.167 \ci{-.486}{+.812} & \\
L3 ($\sim$104k) & 3.431 & 3.549 & +.118 \ci{-.660}{+.889} & H1 not supported \\
\midrule
H2: $\Delta_{\mathrm{L3}}-\Delta_{\mathrm{L1}}$ & & & +.417 \ci{-.438}{+1.312} & Direction only \\
\bottomrule
\end{tabular}
\end{table}

Two honest details qualify the mechanism. First, the counterintuitive strength of the truncated baseline: with a 104k-token haystack, the most recent 32k tokens still score $3.43/5$, because a substantial fraction of LongMemEval-s questions have evidence in the recent window or can be answered parametrically; this compresses the headroom available to any memory system. Second, the question-type decomposition is informative: at L3 the positive contributions concentrate in single-session-assistant ($+1.50$), preference ($+2.00$), and single-session-user ($+0.52$) questions, while temporal-reasoning questions are \emph{negative} ($-0.69$)---consistent with the known weakness that \zifa{}'s summarized L2/L3 representations drop fine-grained temporal detail. We report the negative type as-is rather than excluding it.

\subsection{Backbone boundary cases}
\label{sec:gemini-boundary}

The target claim is deliberately not backbone-agnostic. In the clean Gemini system-comparison cohort, A1 and A5 are positive but uncertain, while A2 content preference and A3 fiction--reality discrimination regress with intervals excluding zero (Table~\ref{tab:gemini}). This boundary case does not negate the observed target-regime gains, but it rules out applying them indiscriminately. The A3 result is also a safety warning: compressing dialogue into declarative memory can remove the framing that a statement was fictional.

\begin{table}[!ht]
\centering
\caption{Independent Gemini system comparison under the DeepSeek-v4 judge. The contaminated historical factorial cohort is not used. The final row is the fresh factorial-cohort A5 memory contrast $M|E-$; its $M-$ cells (2026-07-13) and $M+$ cells (2026-07-10) come from batches three days apart under the same endpoint, judge, and protocol, so it is reported under that disclosed cross-batch qualification (Section~\ref{sec:qualification}).}
\label{tab:gemini}
\small
\begin{tabular}{lrr}
\toprule
Endpoint & $\Delta$ & 95\% CI \\
\midrule
A1 persona & +.079 & \ci{-.009}{+.167} \\
A2 preference (content) & -.061 & \ci{-.118}{-.004} \\
A2 preference (style) & -.029 & \ci{-.085}{+.024} \\
A3 fiction--reality & -.100 & \ci{-.190}{-.010} \\
A5 emotional intelligence & +.141 & \ci{-.106}{+.397} \\
A5 emotional intelligence (fresh factorial cohort, $M|E-$, cross-batch) & .000 & \ci{-.317}{+.321} \\
\bottomrule
\end{tabular}
\end{table}
\FloatBarrier

The new fresh-cohort $M|E-$ row is consistent with the picture that Gemini's raw-history A5 baseline already sits near its ceiling ($4.058$ in both fresh cells), so the structured-context A5 effect on this backbone is small in every admissible cohort: $+0.141$ \ci{-.106}{+.397} in the system-comparison cohort, $0.000$ \ci{-.317}{+.321} in the fresh factorial cohort, and approximately $+0.02$ in the earlier v3.1-judge cohort (directional pattern only; absolute scores are not comparable across judges, Appendix~\ref{app:judge}).

\subsection{Affect injection under memory: null everywhere, and backbone-dependent without it}
\label{sec:factorial}

Once structured memory is present, the legacy affect layer provides no detected incremental benefit on any endpoint. The high-powered A2 endpoints are especially informative: PF-content is $-0.020$ \ci{-0.069}{+0.029}, and PF-style is $+0.004$ \ci{-0.045}{+0.053}. Their upper bounds constrain a missed positive effect to about three and five percentage points under this protocol. A1 is less precise; A3 is ceiling-dominated; and A5 has a wide interval. We therefore do not call all five endpoints equivalent or ``precise nulls.'' The fresh, explicitly routed Gemini A5 replication gives $E|M+=-0.003$ \ci{-0.272}{+0.269}; it does not reproduce the historical negative Gemini cell, but its interval remains too wide to establish equivalence. Other route-clean A5 backbones also have small positive point estimates under $M+$ (Qwen $+0.06$, Doubao $+0.07$, GLM $+0.04$), all with intervals crossing zero.

Without memory, the picture is not a mirror image but a backbone-dependent one. On Claude, the affect block yields modest positive simple effects under raw history (Table~\ref{tab:affect-simple}); on the fresh same-day, same-batch Gemini A5 pair, the same contrast is significantly negative: $E|M-=-0.346$ \ci{-0.696}{-0.032} ($M-E-$ $4.058$ vs.\ $M-E+$ $3.712$). An earlier manuscript reading---that the affect block acts as a compact auxiliary summary whenever structured memory is absent---is therefore corrected: that positive reading is Claude-specific. Across backbones, the only invariant this design supports is that the affect layer is consistently null once memory is present; without memory, its sign varies by backbone. This correction strengthens rather than weakens the paper's central point: an injected affect block is not a reliable channel for emotional continuity.

The attenuation on Claude remains consistent with partial substitution: structured memory already presents the facts, preferences, and emotional history from which the backbone can infer an appropriate tone, so an additional affect summary has little new information to contribute. This interpretation is post-hoc and does not establish the mechanism; it also cannot be symmetric across backbones given the Gemini sign flip. In particular, $E+$ describes companion mood induced by appraisal, not a direct user-affect trace.

\subsection{\dyaffect{}'s latent state fails its primary comparison}
\label{sec:dyaffect-results}

\dyaffect{} is designed to align temporal state with the user's affect rather than the companion's mood. Its strict primary test asks whether latent state estimation adds response value beyond recent affect observations. It does not clear that bar: against C3, the three-observation baseline, C4 has net preference $-0.133$ \ci{-0.317}{+0.050}; it loses consistently on 20 of 60 changed-plan scenarios and wins on 12. Its all-96 estimate is also negative ($-0.083$), and answer-order discordance reaches $26.7\%$. The preregistered advancement criterion is therefore not met, and no confirmatory claim is made. The favorable C4--C2 contrast of Section~\ref{sec:threads} answers a different question---whether multi-turn affect context beats a single snapshot---and does not show that the latent state is valuable: C3 already supplies multi-turn observations and beats C4 on the primary contrast.

\subsection{Mechanical validation explains why a plausible state can still fail}
\label{sec:mechanical-results}

The frozen deterministic suite contains 112 executed trajectories and 51,840 one-second rows. It fails the preregistered analytic and numerical validation gates, so the planned downstream response-quality stage was not run. The largest discrepancy between dense numerical evolution and the claimed analytic fast-forward is $L_\infty=0.335$. Sparse fast-forward applies an entire elapsed interval on one side of a cooldown boundary rather than splitting the interval. Moderate-input saturation is $12.108\%$, above the $5\%$ gate, and 59 of 80 relevant trajectories remain saturated for more than two seconds.

The old memory force also fails its approach criterion. Because the force direction is computed once and held open-loop, the particle can cross the centroid and continue accelerating in the old direction. These failures do not prove that affect dynamics are impossible or that users dislike temporal continuity. They show that mechanical plausibility must be validated before response gains are attributed to a state machine, and that an analytically incorrect fast path can erase the intended behavior.

\FloatBarrier
\section{Discussion}
\label{sec:discussion}

\paragraph{A decision map: when to use what.}
The combined evidence supports workload-conditioned system selection rather than a single winner. If responses must reuse durable relational state---persona, preferences, emotional history---replace raw-history context with a memory system: all three tested memory systems beat raw-history deployment significantly, and on these endpoints raw history is the dominated option. If the workload is short-history preference following where evidence fits a few verbatim excerpts, simple filtered retrieval is a strong, cheap first choice; it preserved exact wording that summarization abstracts away and significantly beat the structured representation on both A2 endpoints. If emotional continuity is the product's center, the A5 contrasts among the three memory systems are not resolved at $n=52$ (the \zifa{} $-$ Mem0 interval spans $-0.445$ to $+0.029$), so the choice must currently be made on integration and operational grounds while better-powered evidence accumulates. For very long histories, the measured gradient favors structured memory directionally but is unproven at $n=48$, and temporal-reasoning questions currently favor representations that retain verbatim, timestamped detail.

\paragraph{The useful unit of claim is a workload, not a universal backbone effect.}
\zifa{} is strongest when the response depends on durable relational state: who the companion is, what the user prefers, and how prior interactions should shape tone. The capability matrix makes that scope operational: gains concentrate where a backbone leaves headroom on the workload, and the targeted projection combines observed persona and preference effects under one explicit, post-hoc workload mix. It is intentionally an application estimate, not evidence that every model or conversation benefits.

\paragraph{Affect information is not the same as affect state.}
Three evidence streams point to this distinction. The legacy block helps some endpoints under raw history on Claude, hurts on Gemini, and is uniformly undetectable under structured memory. The redesigned filter beats a single snapshot yet fails against three recent observations. The deterministic state engine violates several of its own trajectory assumptions. Together, these findings support carrying affect through consolidated, retrievable memory---and treating a latent dynamics layer as an empirical hypothesis, not a default architectural virtue.

\paragraph{Memory changes the marginal value of auxiliary state.}
The attenuation pattern can be understood as information overlap. Structured memory already surfaces preferences, salient events, and an interpretable relationship history. An additional affect summary has less new information to contribute and can compete for prompt attention. The phrase ``affect as a cheap substitute for memory'' is useful shorthand for the Claude pattern, but it remains an interpretation: the experiment did not manipulate information content independently of prompt form, $M-$ contains raw history rather than no history, and the substitute reading does not transfer to Gemini, where the affect block hurt without memory.

\paragraph{Model dependence matters.}
Gemini's A2 and A3 regressions show that context compression can remove information a backbone used successfully from raw history. A memory benchmark that reports only pooled gains can hide this cost. Provenance-preserving summaries, especially explicit fictional-frame tags, are a more direct response to the A3 failure than adding affect state.

\paragraph{What differentiates next.}
Two directions follow directly from the disclosed boundaries. First, update-style workloads: \alps{} Task 2 targets memory revision---preferences that change, facts that get corrected---where verbatim retrieval must reconcile contradictory excerpts and a consolidated user model has a structural advantage that the current endpoints do not measure. Second, temporal representation: the long-history type decomposition (temporal-reasoning $-0.69$) and the C4--C2 result jointly suggest keeping timestamped, verbatim-anchored episodic threads alongside summaries, so that consolidation stops discarding the temporal detail that both retrieval and reasoning need. Both are falsifiable next experiments, not claims.

\section{Related Work}
\label{sec:related}

\paragraph{Memory for LLM agents.}
MemGPT introduced OS-style movement between context tiers \citep{packer2023memgpt}; MemoryBank combined long-term storage with Ebbinghaus-inspired forgetting in a companion setting \citep{zhong2023memorybank}; and Mem0, A-MEM, HippoRAG, and related systems refine extraction, linking, retrieval, and graph organization \citep{chhikara2025mem0,xu2025amem,gutierrez2024hipporag}. LongMemEval and LoCoMo evaluate long-horizon recall and reasoning \citep{wu2025longmemeval,maharana2024locomo}. \zifa{} evaluates how a structured memory representation changes companion responses when the comparator already receives the raw history---a harder bar than an empty-context baseline---and additionally runs a preregistered same-protocol comparison against Mem0 and a filtered verbatim-retrieval baseline (Section~\ref{sec:systems-results}), which finds deployment-level gains for all memory systems and representation-level equivalence on the preregistered primary endpoint rather than a margin over prior work.

\paragraph{Persona and companion response quality.}
PersonaChat, RoleLLM, and CharacterEval study persona-conditioned or role-play behavior \citep{zhang2018personachat,wang2024rolellm,tu2024charactereval}. EmpatheticDialogues and ESConv study emotionally appropriate responses and support strategies \citep{rashkin2019empathetic,liu2021esconv}. \alps{} combines these concerns with persistent memory: the probe asks whether context helps the model use persona, preferences, fictional framing, and emotional history rather than whether it can retrieve a fact in isolation.

\paragraph{Affect representation and dynamics.}
VAD/PAD models represent affect in a continuous valence--arousal--dominance space \citep{russell1980circumplex,mehrabian1996pad}. Appraisal architectures map events to affective consequences \citep{ortony1988occ,marsella2009ema}, while DynAffect and emotional-inertia research motivate temporal state \citep{kuppens2010dynaffect,kuppens2010inertia}. Recent LLM work shows that affective prompts and emotional-support data alter generated behavior \citep{li2023emotionprompt,liu2021esconv}. Our question is comparative: does a latent affect state improve responses beyond the observations and dialogue history from which that state is computed?

\paragraph{LLM judges.}
LLM-as-judge methods enable repeatable response evaluation but exhibit position, style, and model-family biases \citep{zheng2023mtbench,liu2023geval,wang2024fairevaluators}. We use paired comparisons, three judge repeats, route provenance, and an explicit judge bridge. These controls reduce some noise; they do not turn model judgments into human user outcomes.

\section{Limitations and Broader Impact}
\label{sec:limitations}

\paragraph{Target selection and projection.}
The persona- and preference-heavy target regime was chosen after reviewing the outcomes. Its $40/30/30$ workload mix is illustrative rather than estimated from deployment traffic, and its uncertainty envelope combines marginal interval endpoints without modeling cross-endpoint covariance. The projection is useful for scenario planning, but it is not a preregistered effect, a joint confidence interval, or evidence for other backbones.

\paragraph{Comparator scope.}
The core memory comparison is against full raw history under one strong context policy. The preregistered prior-system experiment adds Mem0 and a filtered verbatim-retrieval baseline under an identical protocol, but it covers one backbone (Claude-Sonnet-4.6), two capability endpoints plus A5, and reuses the pilot \zifa{}/raw-history cohorts run five to six days earlier (same endpoint, judge, and prompt; disclosed). MemoryBank, A-MEM, and other published systems were not run, and no claim is made about them.

\paragraph{Long-history power.}
The layered experiment has $n=48$ per layer, and every layer interval crosses zero. Detecting the observed $0.1$--$0.4$ effect magnitudes would require roughly $n\gtrsim150$ per layer or a binarized endpoint. The layers are constructed from LongMemEval-s by seeded filler subsampling; conclusions are conditional on that construction and on the single truncation policy tested (most recent 32k tokens).

\paragraph{Post-hoc factorial analysis.}
The Claude $2\times2$ matrix was completed before this claim structure was chosen. Intervals are paired and the within-layer contrasts are clean, but no multiplicity-adjusted confirmatory test is claimed. The attenuation pattern should be replicated under a newly frozen protocol if it becomes a central causal claim. The Gemini $E|M-$ pair is same-day and same-batch, but it is a single backbone--endpoint cell; the sign-heterogeneity conclusion rests on two backbones.

\paragraph{Authored development suites.}
Both \alps{} Task 4 and the 96-scenario \dyaffect{} suite are selected development benchmarks. \dyaffect{} was iterated on the same scenario family and did not reach confirmation. Results do not estimate effects for individual users, cultures, or deployment traffic.

\paragraph{Judge and order sensitivity.}
All headline judgments use one DeepSeek-v4 route. Three repeats expose nonzero variance, and the \dyaffect{} pairwise test exceeds its order-discordance budget. The judge bridge supports directional robustness for many cells, not judge invariance. Human evaluation of longitudinal companion responses remains necessary.

\paragraph{Construct coverage.}
A1--A5 are response-level proxies. A3 is partly ceiling- or floor-dominated depending on backbone, and A5 has only 52 instances. The mechanical suite tests code trajectories rather than human-perceived believability. No claim is made that the tested dynamics improves well-being, attachment, or emotional regulation.

\paragraph{Broader impact.}
Better memory can improve continuity but also intensify privacy, dependency, manipulation, and false-belief risks in AI companionship \citep{laestadius2022toohuman,pentina2023replika}. A3 demonstrates one concrete harm: summarization may transform role-play into apparent fact. Systems should provide inspectable memory, deletion and correction controls, explicit artificial-identity disclosure, and provenance tags. Affect state should not be used to simulate distress or obligation in ways that pressure a user to remain engaged.

\section{Conclusion}

Emotional continuity is the most general measured gain of structured relational memory: against full raw dialogue history, \zifa{} improves judged emotional intelligence directionally on every verified backbone, with a pooled gain of $+0.42$ \ci{+0.25}{+0.60}, and its affect-tagged consolidation is the mechanism the contrastive evidence favors---multi-turn emotional threads beat single-turn snapshots, while injected affect blocks and latent state machines do not earn their place. Capability-level gains form a targeted map rather than a constant, concentrating where a backbone leaves headroom on persona and preference workloads. The same-protocol comparisons bound the claim honestly: every memory system tested beats raw-history deployment, \zifa{} is statistically equivalent to an existing open system within $\pm5$ percentage points on the preregistered primary endpoint (post-hoc TOST) with emotional-intelligence differences unresolved at $n=52$, and simple filtered verbatim retrieval is a strong baseline for short-history preference tasks; the long-history gradient is directionally positive but unproven at current power. The resulting design principle is targeted and practical---carry emotional history in consolidated, affect-tagged memory; prefer verbatim retrieval where verbatim detail dominates; monitor the backbone boundary regimes; and add dynamics only when it beats information-matched baselines.

\section*{Data and Code Availability}

The public alpha memory SDK is available at \url{https://github.com/zifacorp/zifamem}. \alps{} Task 4 endpoints, the frozen preregistrations for the prior-system and long-history experiments, the row-level route-audit and analysis artifacts, and the \dyaffect{} implementation and analysis materials are archived with this manuscript version for reproducibility. Historical model outputs and judge records contain provider-specific provenance and will be released only where licensing and privacy constraints permit. Test-set gold remains withheld.

\paragraph{Disclosure.}
The reported experiments use public benchmark dialogues curated from real human--LLM conversations (\alps{}, WildChat-derived), an authored affect-transition scenario suite (\dyaffect{}), public long-memory benchmark data, and model-generated responses, not human participants or private deployment conversations. Several authors are affiliated with ZifaCorp, which develops AI companion and memory technology. Language-model tools assisted code analysis and editorial revision; all reported statistics were verified against preserved experiment records or deterministic analysis, and the authors remain responsible for the claims, citations, and final text.

\bibliographystyle{unsrtnat}
\bibliography{references}

\appendix

\section{Judge Bridge}
\label{app:judge}

We re-judged 4,096 rows from the independently routed comparison cohort using the DeepSeek-v4 judge with three repeats. Delta direction is retained in 16 of 20 backbone--endpoint comparisons. The four flips occur in cells whose magnitudes are near the judge-noise band. V4 is systematically stricter on A1/A2 and A5 absolute scores. Estimated 95\% aggregate-delta noise thresholds are approximately $0.042$ for A1, $0.029$ for each A2 endpoint, $0.040$ for A3, and $0.227$ for A5. This bridge supports using v4 consistently within the present analyses; it does not permit absolute-score comparison with the earlier v3.1 table or establish judge invariance. The four-backbone capability matrix of Table~\ref{tab:m-matrix} adds two further directional-consistency examples with the earlier v3.1 cohort (Qwen's A2 friction and its slight A3 regression).

\begin{table}[h]
\centering
\caption{V4 structured-context deltas in the clean four-backbone bridge cohort. Values are point estimates; the paired intervals used for Claude and Gemini are reported in the main text.}
\label{tab:bridge}
\small
\setlength{\tabcolsep}{3pt}
\begin{tabular}{lrrrrr}
\toprule
Backbone & A1 & A2 content & A2 style & A3 & A5 \\
\midrule
Claude-Sonnet-4.6 & +.149 & +.073 & +.089 & +.010 & +.593 \\
Doubao-Seed-2.1-turbo & +.132 & +.020 & +.008 & -.040 & +.349 \\
Gemini-3.5-flash & +.079 & -.061 & -.029 & -.100 & +.141 \\
GLM-5.2 & +.114 & -.008 & .000 & +.070 & +.180 \\
\bottomrule
\end{tabular}
\end{table}

\section{Mechanical Diagnostic Ledger}
\label{app:mechanics}

\begin{table}[h]
\centering
\caption{Selected deterministic gates for the frozen legacy dynamics.}
\label{tab:mechanics}
\small
\begin{tabular}{p{0.28\linewidth}p{0.48\linewidth}l}
\toprule
Check & Observation & Verdict \\
\midrule
Dense vs analytic & Maximum recorded $L_\infty=0.335258$ & Fail \\
Dense vs sparse & Maximum $L_\infty=1.13$; cooldown crossings not segmented & Fail \\
Recovery monotonicity & 12/16 pass; four tsundere trajectories oscillate & Fail \\
Recovery half-life & Corrected errors 29.9\%, 11.8\%, 62.1\%, 12.4\% by preset & Fail \\
Moderate saturation & 5,812/48,000 rows (12.108\%) & Fail \\
Legacy recall force & 14/48 within 10\% at $t=8$; open-loop overshoot observed & Fail \\
Evaluation completeness & 112/112 executions; 51,840 rows; zero parse errors & Pass \\
\bottomrule
\end{tabular}
\end{table}

The legacy mechanics failed the preregistered analytic and numerical validation gates, so the planned downstream response-quality stage was not run. Post-hoc corrections to duplicated events and several denominators did not reverse the underlying failures; they are retained as diagnostic corrections rather than substituted success criteria.

\section{Excluded Evidence and Data Integrity}
\label{app:isolation}

The following evidence items are preserved for audit but excluded from manuscript estimates:
\begin{itemize}
  \item the earlier contaminated Gemini factorial cells and any pooled value containing them;
  \item qualitative ``mood mismatch'' examples selected from those contaminated cells;
  \item superseded emotion-dynamics v2/v3 engineering runs as efficacy evidence; and
  \item incomplete intermediate \dyaffect{} response files.
\end{itemize}
Admitted Gemini evidence is cohort-specific: an independently routed system comparison, a separate replication limited to $E|M+$, and a fresh same-batch factorial pair admitting $E|M-$ plus the disclosed cross-batch $M|E-$ contrast. The completed \dyaffect{} evaluation contains 96 base responses, 194 unique-plan edits, 288 mapped responses, and 196 pairwise judgments; resumed calls remain documented in the audit trail.

\section{Preregistration Status Ledger}
\label{app:prereg-ledger}

Two evidentiary tiers are used in this paper and are never mixed: \emph{preregistered} means a design and stopping rule frozen before the corresponding data were collected or judged; \emph{route-audited exploratory} means no frozen preregistration file exists, but every row carries route and judgment artifacts and the stated stopping discipline was applied.

\begin{table}[h]
\centering
\caption{Evidence streams by preregistration tier.}
\label{tab:prereg-ledger}
\small
\begin{tabular}{>{\raggedright\arraybackslash}p{0.44\linewidth}>{\raggedright\arraybackslash}p{0.50\linewidth}}
\toprule
Evidence stream & Tier \\
\midrule
Prior-system comparison (Section~\ref{sec:exp1-protocol}) & Preregistered, frozen before data (\texttt{positive\_exp\_baselines\_prereg\_20260714}) \\
Long-history layered experiment (Section~\ref{sec:exp2-protocol}) & Preregistered, frozen before data (\texttt{positive\_exp\_longhistory\_prereg\_20260714}) \\
\dyaffect{} contrastive suite & Preregistered gates and caveat rules frozen before judging \\
Legacy dynamics deterministic validation & Preregistered under the versioned emotion-dynamics v2 contract \\
Four-backbone capability matrix (Table~\ref{tab:m-matrix}) & Route-audited exploratory, per-row artifacts; no frozen preregistration \\
Claude $2\times2$ factorial matrix & Route-audited exploratory, per-row artifacts; completed before this claim structure was chosen \\
Judge bridge re-judging (Appendix~\ref{app:judge}) & Route-audited exploratory audit, per-row artifacts \\
TOST equivalence check (Section~\ref{sec:systems-results}) & Post hoc on preregistered per-instance artifacts; margin not prespecified \\
\bottomrule
\end{tabular}
\end{table}

\section{Implementation Provenance}
\label{app:provenance}

\begin{table}[h]
\centering
\caption{Component-level provenance.}
\small
\begin{tabular}{>{\raggedright\arraybackslash}p{0.25\linewidth}p{0.67\linewidth}}
\toprule
Component & Claim boundary \\
\midrule
Public \zifa{} SDK & Alpha memory substrate and examples; not a hosted production service. \\
Internal memory service & Source of the historical structured-context behavior and the corrected promotion/reranking equations. \\
Legacy emotion service/harness & Source of historical $E+$ responses and the frozen mechanical audit; not claimed as the current deployed coupling. \\
\dyaffect{} & Confidence-adaptive user-affect state estimator evaluated through the controlled C2/C3/C4 response contrasts. \\
\alps{} judge harness & Source of response-level endpoints, three-repeat aggregation, documented model routes, and row-level judgments. \\
Prior-system harness & Mem0 and simple-RAG columns generated under the byte-identical Task~4 prompt contract with per-row route records. \\
\bottomrule
\end{tabular}
\end{table}

\end{document}